%% file: main.tex
\documentclass[conference]{IEEEtran}
\IEEEoverridecommandlockouts

\usepackage{cite}
\usepackage{amsmath,amssymb,amsfonts}
\usepackage{algorithmic}
\usepackage{hyperref}
\usepackage{graphicx}
\usepackage{textcomp}
\usepackage{xcolor}
\usepackage{gensymb}
\usepackage{comment}
\usepackage{multirow}  
\usepackage{booktabs}   

\def\BibTeX{{\rm B\kern-.05em{\sc i\kern-.025em b}\kern-.08em
    T\kern-.1667em\lower.7ex\hbox{E}\kern-.125emX}}
\begin{document}

\title{EA: An Event Autoencoder for High-Speed Vision Sensing\\
\thanks{This work was supported in part by the National Science Foundation (NSF) award number: 2138253, the Maryland Industrial Partnerships (MIPS) program under award number MIPS0012, and the UMBC Startup grant.}
}

\author{\IEEEauthorblockN{Riadul Islam, Joey Mulé, Dhandeep Challagundla, Shahmir Rizvi, and Sean Carson}
\IEEEauthorblockA{\textit{Department of Computer Science and Electrical Engineering} \\
\textit{University of Maryland, Baltimore County}\\
Baltimore, Maryland, USA \\
riaduli@umbc.edu, jmule2@umbc.edu, vd58139@umbc.edu, shahmir1@umbc.edu, scarson4@umbc.edu}

}

\maketitle

\begin{abstract}
High-speed vision sensing is essential for real-time perception in applications such as robotics, autonomous vehicles, and industrial automation. Traditional frame-based vision systems suffer from motion blur, high latency, and redundant data processing, limiting their performance in dynamic environments. Event cameras, which capture asynchronous brightness changes at the pixel level, offer a promising alternative but pose challenges in object detection due to sparse and noisy event streams. To address this, we propose an event autoencoder architecture that efficiently compresses and reconstructs event data while preserving critical spatial and temporal features. The proposed model employs convolutional encoding and incorporates adaptive threshold selection and a lightweight classifier to enhance recognition accuracy while reducing computational complexity. Experimental results on the existing Smart Event Face Dataset (SEFD) demonstrate that our approach achieves comparable accuracy to the YOLO-v4 model while utilizing up to $35.5\times$ fewer parameters. Implementations on embedded platforms, including Raspberry Pi 4B and NVIDIA Jetson Nano, show high frame rates ranging from 8 FPS up to 44.8 FPS. The proposed classifier exhibits up to $87.84\times$  better FPS than the state-of-the-art and significantly improves event-based vision performance, making it ideal for low-power, high-speed applications in real-time edge computing.
\end{abstract}


\input{introduction}
\input{background}

\input{proposed_method}

\input{analysis}

\input{conclusion}

\section*{Acknowledgment}

Special thanks
to Sri Ranga Sai Krishna Tummala of UMBC for initial architecture and results, Rohith Kankipati and Ryan
Robucci of UMBC and Chad Howard of Oculi for assisting in the analysis and discussion.

\bibliographystyle{IEEEtran}
\bibliography{main}

\end{document}

%% file: introduction.tex
\section{Introduction}
\label{sec:intro}
High-speed vision sensing is crucial in applications requiring rapid perception and real-time decision-making, such as robotics, autonomous vehicles, industrial automation, and scientific imaging. Traditional frame-based vision systems struggle with motion blur, high latency, and excessive data redundancy when capturing fast-moving scenes. Event cameras capture information similarly to the human eye, where sensors asynchronously detect changes in brightness in individual pixels rather than capturing entire frames on fixed intervals~\cite{PROPHESEE:2024, Joey_ebcd:2025}.

This results in ultra-fast response times, reduced data bandwidth, and energy-efficient processing, making event vision particularly advantageous for low-latency, high-dynamic-range, and power-constrained environments. By capturing only relevant motion information, event-based sensors enable more efficient tracking, object detection, and recognition in dynamic and high-speed scenarios, overcoming the limitations of conventional vision systems. However, building an object detection model for event sensing is non-trivial due to the lack of information. Hence, this research proposed an event autoencoder model for low-cost high-speed operation.

Autoencoders are highly effective for high-noise object detection because they learn compact, denoised representations of input data by capturing essential features while filtering out irrelevant noise~\cite{Wang_auto:2016, Tai_pottsmgnet:2024, gruel2022event}. Their ability to reconstruct clean data from noisy inputs makes them ideal for event-based vision, where raw event streams are often sparse and contain noise. By encoding spatiotemporal patterns in event data, autoencoders can enhance signal quality, improve feature extraction, and facilitate robust object detection in challenging environments. 


Researchers applied spatiotemporal transformers for streaming object detection~\cite{li2023sodformer}. However, this optical frame flow-based approach is slow and has a low mean average precision (mAP). Another interesting work generates a binary event history image (BEHI) using an event encoder and a simple four-state convolutional neural network (CNN) for high-speed object detection~\cite{wang2022ev}. However, this approach requires additional uncertainty measurement techniques for motion estimation. Besides, developing a scalable architecture tailored for adaptive event vision necessitates comprehensive research. 


Unlike time-dependent spiking neural networks (SNNs)~\cite{islam2024benchmarking, comcsa2021spiking, kamata2022fully, islam2023exploring}, this research introduces a novel static event autoencoder-based vision classification architecture to address these limitations.
The architecture specializes in event-based detections, enabling an adaptive threshold selection mechanism for energy-efficient computer vision tasks with event cameras. Adaptive threshold selection dynamically adjusts the sensitivity of event detection based on environmental conditions and signal characteristics, effectively reducing redundant event generation while preserving essential visual information. In particular, the main contributions of this research are
\begin{itemize} \renewcommand{\labelitemi}{$\bullet$}
		
	   \item We present a new event autoencoder architecture for event-based face detection.
       \item We verified the scalability and robustness of the architecture considering different threshold event data.
          \item We analyzed the effectiveness of the proposed architectures on embedded platforms like Raspberry Pi and NVIDIA Jetson Nano boards.
	   \item The proposed event classifier has $35.47\times$ lower parameters and up to $21.44\times$ faster frames per second (FPS) compared to the YOLO-v4~\cite{Bochkovskiy_yolov4:2020} model.
    \end{itemize}



%% file: background.tex
\section{Background}
\label{sec:background}
Deep learning-based autoencoder exhibits promising results in resolving sparsity issues in multi-modal recommendation systems~\cite{rajput2024autoencoder}. Researchers also use CNN-based autoencoder for the medical image compression and classification~\cite{fettah2024convolutional}. 

Other researchers have explored the use of recurrent neural network (RNN)-based autoencoders for optical flow prediction in event-based vision~\cite{Wan_event_auto:2022, Li_rnn_auto:2020, Gehrig_rnn_auto:2021, Hidalgo_rnn_auto:2020, Yu_spikingvit:2025, li2022asynchronous, kim2022ev, zhang2022spiking}. These models leverage the sequential processing capabilities of RNNs to capture the temporal dependencies in event streams, making them well-suited for motion estimation and tracking tasks. Optical flow prediction is crucial in event-based vision, as it enables accurate motion analysis in dynamic environments by estimating the movement of objects based on sparse and asynchronous event data. Other researchers used multiple encoders for event-fusion~\cite{Han_auto_fusion:2020, Zhang_auto_fusion:2021}.


However, despite their advantages, these approaches may not be suitable for low-cost and high-speed event sensor design. The computational complexity and high power consumption of RNN-based architectures and multi-encoder frameworks pose significant challenges for resource-constrained applications, such as edge computing and real-time embedded systems. Compared to other energy-efficient computation techniques~\cite{islam_dcmcs:2018, Kodukula_dvfs:2023, Challagundla_tvlsi:2024, islam2011high}, implementing these models on low-power event cameras may lead to increased latency and energy inefficiency, limiting their practical applicability in real-world scenarios where power efficiency is a critical factor.
To address these limitations, this research introduces a novel scalable architecture that would enable adaptive threshold selection for energy-efficient computer vision tasks in event cameras. 

%% file: proposed_method.tex
\section{Proposed Methodology}
\begin{figure*}[t!]
		\begin{center}
			\vspace{-0.7cm}
			\includegraphics[width = 0.95\textwidth]{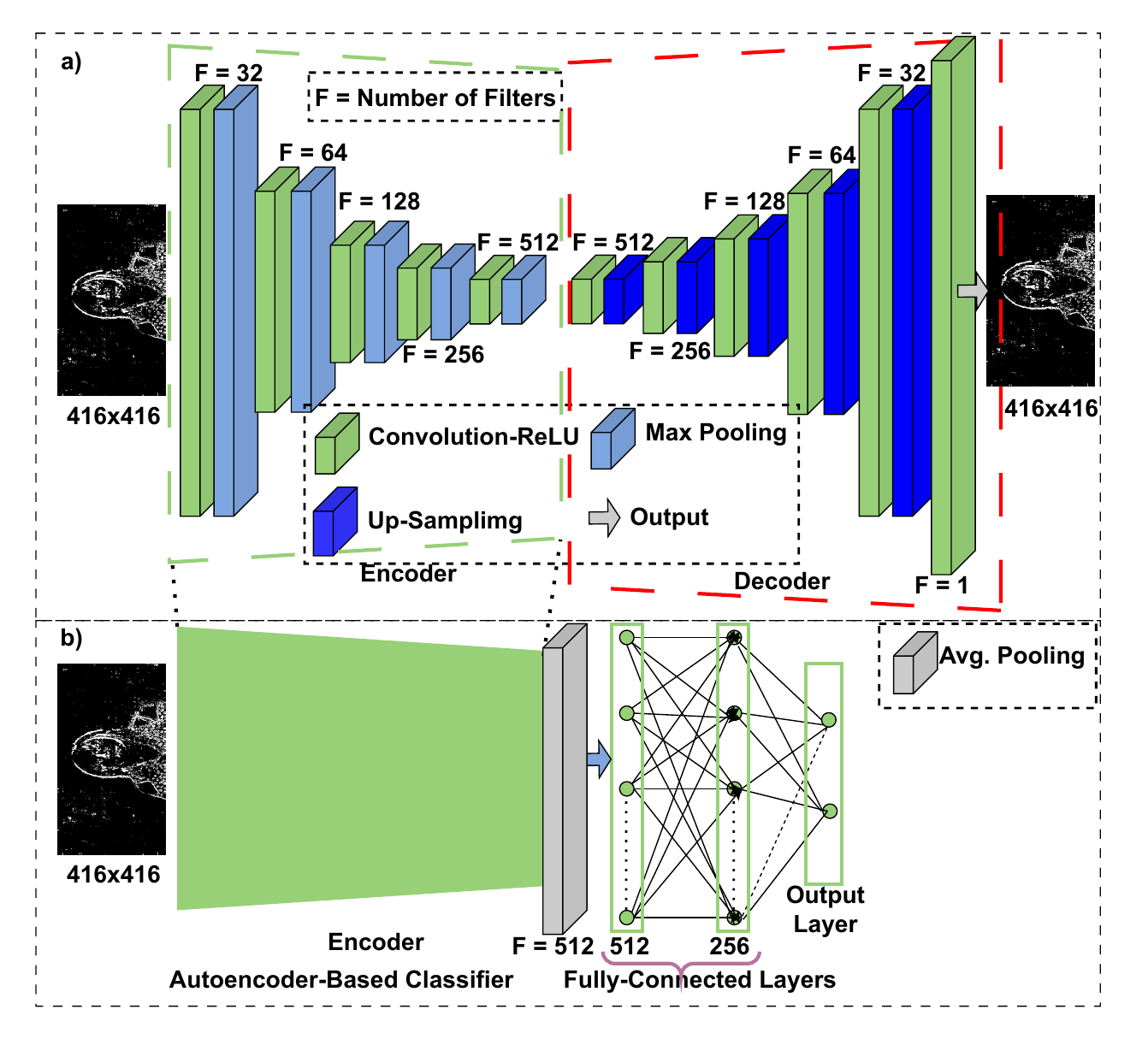}
			\vspace{-0.7cm}
			\caption {(a) The proposed event autoencoder architecture, and (b) proposed autoencoder-based event-classifier.
				\label{fig:autoencoder}}
			\vspace{-0.05cm}
		\end{center}
\end{figure*}
Autoencoders are robust neural network architectures designed to learn efficient, lower-dimensional representations of input data while retaining essential features~\cite{Wang_auto:2016, Tai_pottsmgnet:2024, gruel2022event}. They achieve this by compressing high-dimensional data into a latent space representation through an encoder and then reconstructing the original input using a decoder. This ability to reduce dimensionality while preserving key characteristics makes autoencoders highly useful in applications such as image compression, anomaly detection, and feature extraction.
In various domains, especially those dealing with large-scale or high-dimensional data, dimensionality reduction is crucial for improving the speed of operation, reducing storage requirements, and enhancing the performance of downstream tasks. 

A key advantage of autoencoders is that they operate in an unsupervised manner, meaning they do not require labeled data for training~\cite{Chen_auto_cvpr:2023}. Instead, they learn representations by attempting to reconstruct their own input. This is particularly beneficial in scenarios where labeled data is scarce or expensive to obtain, making autoencoders highly suitable for unsupervised feature learning and representation learning.

From an information theory perspective, autoencoders exhibit data processing inequalities—a fundamental property that describes how information is transformed as it passes through layers of a neural network. Specifically, the mutual information between the input data $X$ and intermediate latent representations $T_i$ follows the relationship:
\begin{equation}
    I(X, T_1) \geq I(X, T_2) \geq \dots I(X, T_L)
\end{equation}
Similarly, for the decoder, the mutual information between the reconstructed input $X'$ and the corresponding latent representations $T_i'$ follows:
\begin{equation}
I(X', T_1') \geq I(X', T_2') \geq \dots I(X', T_L')
\end{equation}
where $I(X, T_i)$ is the mutual information of $i^{th}$ layer neural network, $(X, T_i)$ is represents encoder and $(X', T_i')$ decoder mutual information~\cite{Shujian_Iauto:2018}.

The proposed event autoencoder architecture is designed to process event-based vision data efficiently, as illustrated in Figure~\ref{fig:autoencoder}(a). The autoencoder consists of two main components, an encoder and a decoder, which work together to learn compact representations of event data and reconstruct the original input.

The encoder is responsible for extracting meaningful features from the input event stream. It comprises multiple convolutional layers, each followed by the Rectified Linear Unit (ReLU) activation function and Max pooling layers. 
The convolutional layers apply trainable filters to the input data and preserve critical structural details of the event stream. 
The ReLU activation function introduces sparsity and computational efficiency by setting negative values to zero while allowing positive values to pass through. ReLU activation enhances the learning process by preventing the vanishing gradient problem, enabling the network to capture complex mappings between the input and latent space efficiently. The Max Pooling layers reduce the spatial dimensions of feature maps by selecting the maximum value within a local pooling window. This operation retains the most salient local features while discarding less important information, thereby improving computational efficiency. Max pooling also introduces translational invariance, ensuring that small shifts in the input data do not significantly affect feature extraction.

The decoder complements the encoder by reversing the dimensionality reduction process and reconstructing the input event data. Instead of Max pooling layers, it employs Up-Sampling layers to gradually restore the spatial resolution. The Up-Sampling layers increase the spatial dimensions of the feature maps by interpolating neighboring pixel values. This operation helps recover the fine-grained details of the input data, ensuring that the reconstructed output closely resembles the original event representation.
The decoder performs contextual information integration by progressively increasing spatial resolution, and the Up-Sampling layers allow the autoencoder to capture long-range dependencies in the input data. This is crucial for event-based vision, where sparse and asynchronous event streams require efficient reconstruction techniques to maintain spatial coherence.

During training, both the encoder and decoder architectures are optimized together using a loss function (e.g., Mean Squared Error or Binary Cross-Entropy). The encoder compresses the input into a compact latent representation while the decoder attempts to reconstruct the original event data from this representation. By minimizing reconstruction error, the event--autoencoder learns to extract high-level features that are useful for downstream tasks.

To transform our trained autoencoder model into a classifier, we introduced two fully connected (FC) layers while keeping the encoder layers frozen during the training. This approach leverages the feature extraction capabilities of the pre-trained encoder while allowing the newly added FC layers to learn classification-specific patterns. The overall process is illustrated in Figure~\ref{fig:autoencoder}(b). The classifier design enables the model to efficiently classify event-based data by leveraging the strengths of the pre-trained autoencoder, making it well-suited for real-time event recognition, object detection, and other event-driven vision tasks in low-power and edge-computing applications.

%% file: analysis.tex
\section{Results and Analysis}

{\bf Experimental Setup:} We used the existing Smart Event Face Dataset (SEFD)~\cite{Islam_descriptor:2024, islam2024descriptorfacedetectiondataset}, which was originally generated from the Aff-Wild database~\cite{Kollias:2019}, consisting of 298 videos, ranging from 0.1 to 14.47 minutes. 

\begin{figure}[t]
\begin{center}
\vspace{-.7cm}
\includegraphics[width = 0.5\textwidth]{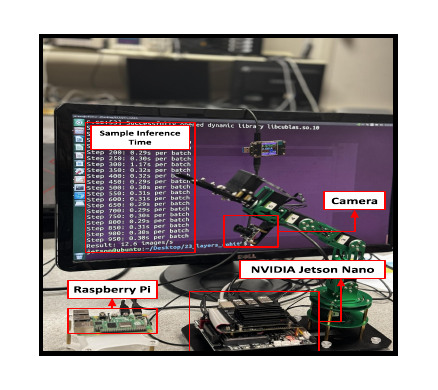}
\vspace{-1.5cm}
\caption[ ]
{ Hardware setup with Raspberry Pi 4B and NVIDIA Jetson Nano boards for proposed models' performance analysis.}
\label{fig:hw_setup}
\vspace{-0.850cm}
\end{center}
\end{figure}

For our initial experiments, we trained state-of-the-art (SOTA) neural network models, including YOLO-v4~\cite{Bochkovskiy_yolov4:2020} and YOLO-v7~\cite{Wang_yolov7:2022}, EfficientDet-b0~\cite{Tan_efficientdet:2020} and MobileNets-v1~\cite{Howard_mobilenets:2017}). These models are characterized by their lightweight nature and effectiveness in object detection tasks. In this research, we considered four thresholds ($T_h = 4, 8, 12, 16$), which are the measure of the difference in pixel intensity between successive image frames or the difference between the pixel intensity of an image frame and a reference frame. For SOTA model training and testing, we used the SEFD dataset. For the proposed event autoencoder-based classifier, we also used the SEFD dataset as well as an equal amount of negative images (i.e., images without faces). The distribution of training and testing subsets are shown in Table~\ref{tab:benchmarks} under the ``Dataset split" columns. All the analysis was performed on an Intel Xeon(R) 20-core CPU with 32 GB RAM and a 4 GB NVIDIA T-1000 Graphics card running Ubuntu 20.04.6 LTS. The analyzed results include commonly referenced evaluation metrics of machine learning (ML): Accuracy, Precision, Recall, and F1 Score. 
To exhibit the effectiveness of the proposed model on embedded platforms, we deployed the event autoencoder model on Raspberry Pi 4B and NVIDIA Jetson Nano boards, the hardware setup is shown in Figure~\ref{fig:hw_setup}.

\begin{table*}[t!]
\vspace{-0.15cm}
	\renewcommand{\arraystretch}{1.2}
	\caption{Compared to the existing SOTA models, the proposed autoencoder classifier expresses greater accuracy percentages than YOLO-v4, YOLO-v7, and MobileNets-v1. While EfficientDet-b0 maintains the best accuracy metrics, the proposed classifier has $2.3\times$ and $35.47\times$ fewer parameters than EfficientDet-b0 and YOLO-v4 models, respectively.
		 \label{tab:benchmarks}}
    \centering
    \vspace{-0.150cm}
    \begin{tabular}{|c|c|c|c|c|c|c|c|c|c|c|}
    \hline
        \multirow{2}{*}{ Models} &\multirow{2}{*}{\# of Parameters} & \multirow{2}{*}{$T_h$} & \multicolumn{2}{c|}{Dataset split}  &  \multicolumn{4}{c|}{Metrics}  \tabularnewline
	
			\cline{4-9}
        ~ & ~ & ~ & Training & Testing  & Accuracy & Precision & Recall & F1-score  \\ \hline
        \multirow{4}{*}{YOLO-v4~\cite{Bochkovskiy_yolov4:2020}} &  \multirow{4}{*}{60.3M} & 4 & 6392 & 904 & 83.81   & 94.00 & 89.00 & 91.00  \tabularnewline 
			\cline{3-9}
        ~ & ~ & 8 & 6392 & 904 & 80.33  & 89.00 & 89.00 & 89.00   \tabularnewline 
			\cline{3-9}
        ~ & ~ & 12 & 6392 & 904 & 79.98 & 91.00 & 86.00 & 89.00  \tabularnewline 
			\cline{3-9}
        ~ & ~ & 16 & 6392 & 904 & 74.49 & 93.00 & 79.00 & 85.00  \\ \hline
         \multirow{4}{*}{YOLO-v7~\cite{Wang_yolov7:2022}} &  \multirow{4}{*}{25.2M} & 4 & 6392 & 904 & 86.71  & 94.00 & 92.00 & 93.00  \tabularnewline 
			\cline{3-9}
        ~ & ~ & 8 & 6392 & 904 & 82.95  & 92.00 & 90.00 & 91.00   \tabularnewline 
			\cline{3-9}
        ~ & ~ & 12 & 6392 & 904 & 83.44 & 95.00 & 87.00 & 91.00   \tabularnewline 
			\cline{3-9}
        ~ & ~ & 16 & 6392 & 904 & 81.88 & 95.00 & 85.00 & 90.00  \\ \hline
         \multirow{4}{*}{EfficientDet-b0~\cite{Tan_efficientdet:2020}} &  \multirow{4}{*}{3.9M} & 4 & 6392 & 904 & 94.55 & 97.68 & 96.72 & 97.21   \tabularnewline 
			\cline{3-9}
        ~ & ~ & 8 & 6392 & 904 & 93.62 & 97.13 & 96.28 & 96.71  \tabularnewline   
			\cline{3-9}
        ~ & ~ & 12 & 6392 & 904  & 97.34 & 97.34 & 95.96 & 96.64   \tabularnewline 
			\cline{3-9}
        ~ & ~ & 16 & 6392 & 904 & 97.30 & 97.30 & 94.54 & 95.91   \\ \hline
          \multirow{4}{*}{MobileNets-v1~\cite{Howard_mobilenets:2017}} &  \multirow{4}{*}{6.05M} & 4 & 6392 & 904 & 91.64 & 96.65 & 94.64 & 95.64   \tabularnewline 
			\cline{3-9}
        ~ & ~ & 8 & 6392 & 904 & 89.44 & 92.64 & 96.28 & 94.43   \tabularnewline 
			\cline{3-9}
        ~ & ~ & 12 & 6392 & 904 & 92.11 & 96.36 & 95.41 & 95.88  \tabularnewline 
			\cline{3-9}
        ~ & ~ & 16 & 6392 & 904  & 89.11 & 95.61 & 92.91 & 94.24   \\ \hline
        \multirow{4}{*}{{\textbf{This work: Autoencoder classifier}}} 
        & \multirow{4}{*}{\textbf{1.7M}} & \textbf{4}  & \textbf{43492} & \textbf{9780} & \textbf{91.82} & \textbf{89.07} & \textbf{89.07} & \textbf{89.07} \tabularnewline 
			\cline{3-9}
         & ~ & \textbf{8}  & \textbf{43492} & \textbf{9780} & \textbf{93.01} & \textbf{88.27} & \textbf{93.77} & \textbf{90.94} \tabularnewline 
			\cline{3-9}  
        & ~ & \textbf{12} & \textbf{43492} & \textbf{9780} & \textbf{92.15} & \textbf{86.70} & \textbf{93.33} & \textbf{89.89} \tabularnewline 
			\cline{3-9}  
         & ~ & \textbf{16} & \textbf{43492} & \textbf{9780} & \textbf{90.02} & \textbf{82.80} & \textbf{92.57} & \textbf{87.41} \tabularnewline 
			\cline{3-9}  
\hline

    \end{tabular}
    \vspace{-0.20cm}
\end{table*}

\begin{figure}[t]
\begin{center}
\includegraphics[width = 0.5\textwidth]{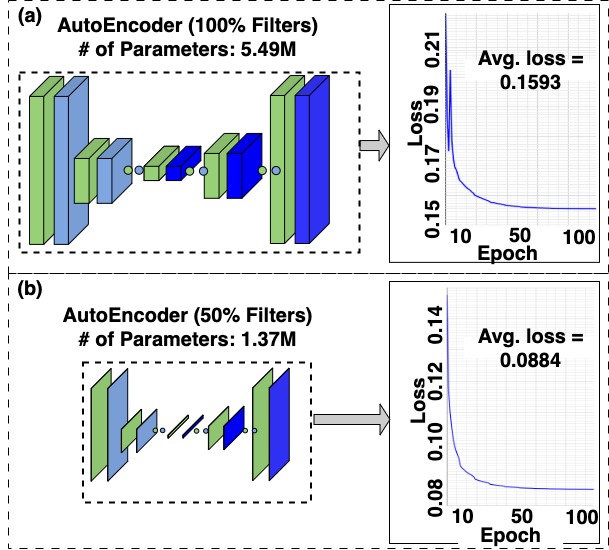}
\vspace{-0.7cm}
\caption[ ]
{(a) The proposed event autoencoder has an average training loss of 0.1593, (b) We demonstrated the robustness of the proposed architecture, considering the 50\% filter model, which has an average training loss of 0.0884.}
\label{fig:loss_autoencoder}
\vspace{-0.7cm}
\end{center}
\end{figure}

{\bf Results and Discussion:} The proposed event autoencoder used both the positive (i.e., SEFD event) and negative (i.e., no face) datasets. The proposed event autoencoder achieves an average training loss of 0.1593, as illustrated in Figure~\ref{fig:loss_autoencoder}(a). When utilizing only 50\% of the filters, the average training loss decreases to 0.0884, as shown in Figure~\ref{fig:loss_autoencoder}(b). In contrast to conventional performance enhancement approaches~\cite{islam2018negative, fung2016dynamics, Islam_asicon:2011, taladriz2025emep, franch2007chip, islam2015differential, moussa2024minimizing}, this research demonstrates the effectiveness of the proposed event-based autoencoder by evaluating its performance under a 50\% reduction in filter size. Reducing the model from 100\% to 50\% of the filters results in a fourfold reduction in model size, demonstrating the robustness of the proposed model in optimizing computational efficiency while maintaining performance.

\begin{figure*}[t!]
		\begin{center}
			\vspace{-0.250cm}
			\includegraphics[width = 0.95\textwidth]{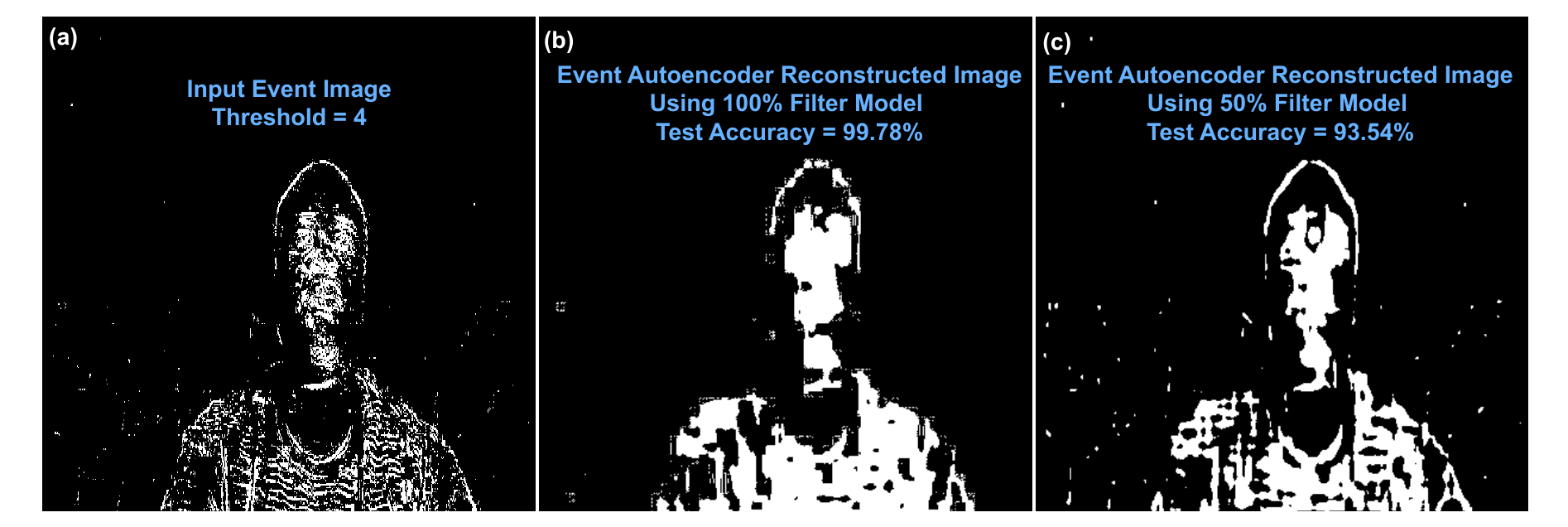}
			\vspace{-0.2cm}
			\caption {The reconstruction test results exhibit excellent performance of the proposed model, (a) a sample event image, (b) the  100\% model shows the average reconstruction test accuracy of 99.97\%, and (c) the scaling robustness of the proposed event autoencoder with 50\% filters has an average test accuracy of in terms of reconstruction test accuracy of 93.54\%.
				\label{fig:autoencoder_test}}
			\vspace{-0.35cm}
		\end{center}
\end{figure*}
The reconstruction accuracy of CNN-based autoencoders is critical for applications such as feature extraction, anomaly detection, denoising, and medical imaging. High accuracy ensures that essential data features are preserved, improving downstream tasks like classification and segmentation. In anomaly detection, better reconstruction helps distinguish normal patterns from outliers, reducing false positives and negatives.
Figure~\ref{fig:autoencoder_test} shows the efficacy of the proposed event autoencoder model. In this example, an input event image of the autoencoder is shown in Figure~\ref{fig:autoencoder_test}(a). The 100\% filter model exhibits a test accuracy of 99.97\%, as shown in Figure~\ref{fig:autoencoder_test}(b). The 50\% model exhibits excellent reconstruction accuracy of 93.54\% while reducing the overall model size by four times, as shown in Figure~\ref{fig:autoencoder_test}(c).

\begin{figure*}[t]
		\begin{center}
			\vspace{-0.7cm}
			\includegraphics[width = 0.95\textwidth]{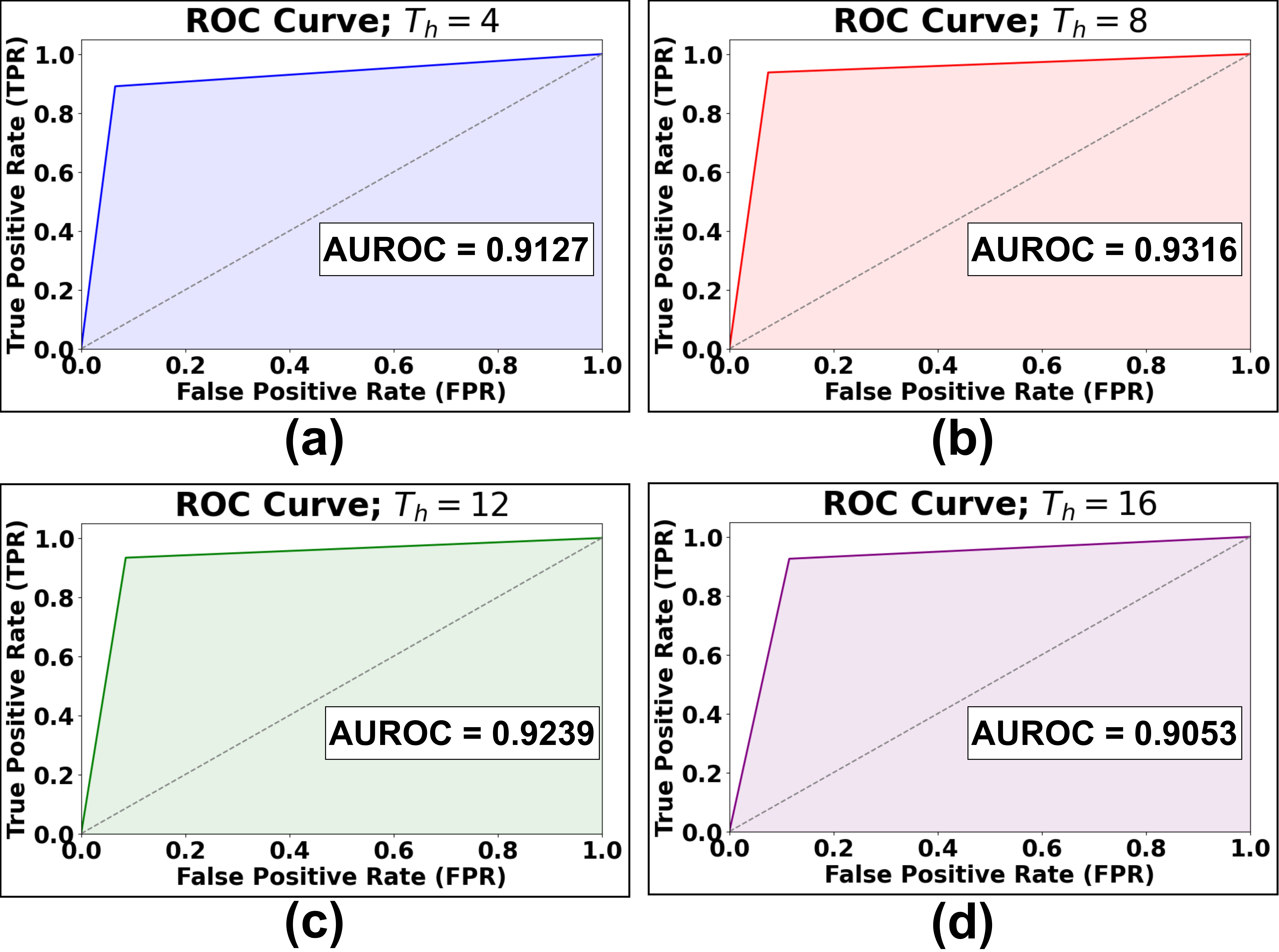}
			\vspace{-0.4cm}
			\caption {The ROC Curve of the proposed autoencoder at different $T_h$ valued above 90\%, in particular, (a) Using $T_h = 4$ has a moderate value of 91.27\%, (b) $T_h = 8$ returns the highest area of 93.16\%, (c) $T_h = 12$ has a moderate value of 92.39\%, and (d) $T_h = 12$ returns the lowest area of 90.53\%.}
				\label{fig:auroc}
			\vspace{-0.15cm}
		\end{center}
\end{figure*}

Evaluating the performance of binary classifiers also commonly involves the receiver operating characteristic (ROC) curve, a critical tool within ML and diagnostic classification tasks. The curve graphically represents the trade-off between the true positive rate (TPR) and the false positive rate (FPR). This graphical representation illustrates how classifiers distinguish between positive and negative instances and is instrumental in understanding model performance across different decision thresholds~\cite{Fawcett:2006}. The area under the ROC curve (AUROC) summarizes this performance to a numerical value, where an area of one indicates perfect classification, and 0.5 corresponds to performance equivalent to random chance~\cite{Hanley:1982}. AUROC is thus widely adopted in various fields, including anomaly detection and out-of-distribution (OOD) detection, to compare and optimize classifier effectiveness. Figure~\ref{fig:auroc} highlights the AUROC performance of the autoencoder classifier across all thresholds, $T_h$. The classifier achieves strong discriminative capability at the lowest threshold ($T_h = 4$), distinguishing between positive and negative events even at high pixel activity representation. Increasing the threshold to $T_h = 8$ yields optimal performance, reflected by the highest AUROC score.
Further increases in threshold values ($T_h = {12}$ and $T_h = {16}$) decrease the AUROC slightly, but still withholding acceptable values on sparse pixel intensity.

{\bf Performance on Embedded Platforms:} The proposed event autoencoder running on a Raspberry Pi 4B board with 50\% filters has $2.71\times$ and $55.24\times$ higher FPS compared to the 100\% model and MobileNets-v1~\cite{Howard_mobilenets:2017} model, respectively. On the same Raspberry Pi board, the event autoencoder-based classifier exhibits $2.07\times$ and $98.57\times$ higher FPS compared to the 100\% model and MobileNets-v1~\cite{Howard_mobilenets:2017} model, respectively. Besides, our classifier with 50\% filter model has $13.44\times$ higher FPS than the EfficientDet-b0~\cite{Tan_efficientdet:2020} model.
As mentioned, this research also characterized the proposed model using the NVIDIA Jetson Nano hardware. As detailed in Table \ref{tab:hardware_benchmarks}, the proposed event autoencoder with 50\% filters exhibits $8.95\times$ and $36.67\times$  better FPS than YOLO-v4~\cite{Bochkovskiy_yolov4:2020} and MobileNets-v1~\cite{Howard_mobilenets:2017}, respectively. The proposed event classifier with 50\% filters exhibits $21.44\times$ and $87.84\times$  better FPS than YOLO-v4~\cite{Bochkovskiy_yolov4:2020} and MobileNets-v1~\cite{Howard_mobilenets:2017}, respectively.

\begin{table}[]
\centering
\renewcommand{\arraystretch}{1.2}
\caption{A comparison of the proposed Autoencoder and classifier runtime with existing neural architectures implemented on Raspberry Pi and Jetson Nano hardware shows significantly lower runtime with considerable increases in FPS in both reconstructing and inferencing images.}
\label{tab:hardware_benchmarks}
\resizebox{\columnwidth}{!}{%
\begin{tabular}{|c|c|c|c|}
\hline
\textbf{Model} & \textbf{Model Size} & \begin{tabular}[c]{@{}c@{}}Raspberry PI\\ FPS (images/s)\end{tabular} & \begin{tabular}[c]{@{}c@{}}Jetson Nano\\ FPS (images/s)\end{tabular} \\ \hline
YOLO-v4~\cite{Bochkovskiy_yolov4:2020} & Full & 0.058 & 2.09  \\ \hline
YOLO-v7~\cite{Wang_yolov7:2022} & Full & 0.08 & 2.73 \\ \hline
EfficientDet-b0~\cite{Tan_efficientdet:2020} & Full & 1.54 & 5.41 \\ \hline
MobileNets-v1~\cite{Howard_mobilenets:2017} & Full & 0.21 & 0.51 \\ \hline
\multirow{2}{*}{\textbf{This work: Autoencoder}} & \textbf{Full} & \textbf{5.6} & \textbf{8.0} \\ \cline{2-4} 
 & \textbf{50\%} & \textbf{11.6} & \textbf{18.7} \\ \hline
\multirow{2}{*}{\textbf{This work: Classifier}} & \textbf{Full} & \textbf{10.0} & \textbf{21.2} \\ \cline{2-4} 
 & \textbf{50\%} & \textbf{20.7} & \textbf{44.8} \\ \hline
\end{tabular}%
}
\end{table}

%% file: conclusion.tex
\section{Conclusion}




This work presents a scalable event autoencoder and classifier aimed at enhancing high-speed vision sensing while minimizing computational overhead. Our proposed method effectively compresses and reconstructs event-based visual information, preserving critical spatial and temporal features. Experimental results on the existing SEFD dataset demonstrate that our approach achieves accuracy comparable to SOTA classifiers while utilizing significantly fewer parameters, making it well-suited for low-power edge computing applications. Furthermore, implementations on embedded platforms such as the Raspberry Pi 4B and NVIDIA Jetson Nano highlight the real-time efficiency of our classifier, achieving $8.28\times$, $21.44\times$, and $87.84\times$  better FPS than EfficientDet-b0~\cite{Tan_efficientdet:2020}, YOLO-v4~\cite{Bochkovskiy_yolov4:2020}, and MobileNets-v1~\cite{Howard_mobilenets:2017}, respectively.
